\documentclass[conference]{IEEEtran}
\IEEEoverridecommandlockouts
\usepackage{cite}
\usepackage{amsmath,amssymb,amsfonts}
\usepackage{algorithmic}
\usepackage{graphicx}
\usepackage{textcomp}
\usepackage{xcolor}
\usepackage{hyperref}
\hypersetup{hidelinks}
\usepackage{url}
\usepackage{tikz}
\usepackage{pbalance}
\usepackage{flushend}
\usepackage{dblfloatfix}
\usepackage{multirow}
\usepackage{multicol}
\usepackage{subcaption}
\usetikzlibrary{shapes,arrows}
\def\BibTeX{{\rm B\kern-.05em{\sc i\kern-.025em b}\kern-.08em
    T\kern-.1667em\lower.7ex\hbox{E}\kern-.125emX}}
\begin{document}

\title{AI Content Self-Detection for Transformer-based Large Language Models\\}

\author{
    \IEEEauthorblockN{Antônio Junior Alves Caiado and Michael Hahsler}
    \IEEEauthorblockA
    {
        \textit{Deptartment of Computer Science} \\
        \textit{Lyle School of Engineering, Southern Methodist University}\\
        Dallas, Texas \\
        acaiado@smu.edu
    }
}

\maketitle

\begin{abstract}
    The usage of generative artificial intelligence (AI) tools based on large language models, including ChatGPT, Bard, and Claude, for text generation has many exciting applications with the potential for phenomenal 
    productivity gains. One issue is authorship attribution when using AI tools. This is especially 
    important in an academic setting where the inappropriate use of generative AI tools may hinder student 
    learning or stifle research by creating a large amount of automatically generated derivative work. 
    Existing plagiarism detection systems can trace the source of submitted text but are not yet equipped with methods to accurately detect AI-generated text. This paper introduces the idea of direct origin detection and evaluates whether generative AI systems can recognize their output and 
    distinguish it from human-written texts. We argue why current transformer-based models may be able
    to self-detect their own generated text and perform a small empirical study using zero-shot learning 
    to investigate if that is the case.
    Results reveal varying capabilities of AI systems to identify their generated text.
    Google's Bard model exhibits the largest capability of self-detection with an accuracy of 94\%,
    followed by OpenAI's ChatGPT with 83\%. On the other hand, Anthropic's Claude model seems to be not
    able to self-detect.\\
\end{abstract}

\section{Introduction}
    Generative AI models like the large language models ChatGPT have become very popular and are applied
    for many tasks that involve question answering and generating text in general. 
    Many exciting applications show the potential for phenomenal productivity gains.
    Such tasks include text summarization, generating explanations,
    and answering questions in general.  
    These models aim to complete a text prompt by mimicking a human-like response. 
    
    While legitimate use of the see models increases, inappropriate use also grows. This is an
    issue in many areas and especially problematic in cases of academic dishonesty where 
    AI-generated text is presented as authentic intellectual work by students or researchers.
    As the models get closer to producing human-like text, detecting AI-generated text becomes increasingly challenging \cite{uchendu-etal-2020-authorship}. Conventional plagiarism detection methods rely on 
    text similarity with known sources, which is inadequate for identifying AI-generated content that represents a new, paraphrased, and integrated version of multiple sources. This limitation requires profound reconsideration of what constitutes plagiarism in the age of generative AI.
    
    Recent research works have explored novel approaches to identify the specific source AI system responsible for text generation rather than detecting plagiarism only. These approaches 
    try to identify artifacts produced by the generation process and
    range from analyzing statistical patterns to stylistic cues\cite{arase2021style} and authorship attribution\cite{uchendu-etal-2020-authorship}. However, even these detection techniques encounter difficulties when AI-generated content is paraphrased or modified to disguise its origins.
    
    This paper proposes a novel method for origin detection called self-detection. It involves using a generative AI system's capability to distinguish between 
    its own output and human-written texts. We will argue why current transformer-based language models 
    should be able to identify their own work. Then, we will use a set of controlled text samples to assess if leading language models, such as OpenAI's ChatGPT, Google's Bard, and Anthropic's Claude can accurately detect their own output. The results demonstrate the limitations of self-detection by AI systems and the potential for evading plagiarism checks through paraphrasing techniques. These findings emphasize the need to reconsider plagiarism and develop more robust techniques for identifying AI-generated content.

    To summarise, the contribution of this paper is as follows:
    \begin{itemize}
        \item We address the struggle of plagiarism detection methods to identify text generated using AI tools.
        \item We propose the novel idea of self-detection, where the tool itself is used to detect AI-generated text.
        \item We provide a small study to examine the ability of AI systems to differentiate between human-written and 
        AI-generated text. 
    \end{itemize}

    This paper first summarizes the background and discusses related 
    studies. We then introduce self-detection and discuss why 
    transformer-based models should have the capability of 
    detecting their own generated text, and we describe several hypotheses. In the experiments section, we evaluate the hypotheses.
    The paper closes with a discussion of the findings.
\section{Background}
    \subsection{Generative AI}
        Generative models are statistical models that 
        learn the joint probability distribution of the data-generating process. Such models
        are often used in machine learning for classification tasks~\cite{mitchell1997machine}, but they can also generate new data following their model. 
        The research of generative models in AI~\cite{russell2010artificial} accelerated after the invention of 
        Variational Autoencoders~(VAE)\cite{kingma2022autoencoding} in 2013, and
        Generative Adversarial Networks~(GAN)\cite{NIPS_GANs} 
        in 2014. A milestone for text data was the development of the transformer 
        architecture\cite{transformer_paper} 
        which is the basis for models, including 
        OpenAI's family of generative pretrained transformers~(GPT)\cite{radford2018improving}
        and other large language models. This technology enables 
        the capability to produce realistic human-like text. An offspring of GPT, ChatGPT pushed the boundaries of natural text generation, enabling the capabilities to produce contextually relevant text. 

        Generative AI is also used to create other types of content
        including images, but we focus on text only in this paper.

    \subsection{Detection of AI-generated Text}
        While detection of AI-generated content can be important in many
        settings, 
        the emergence of generative AI creates especially complicated ethical challenges for academic integrity. Much work
        had already been done to detect plagiarism, which can 
        lead to students not learning by copying assignment solutions
        or researchers taking credit for someone else's work and ideas. 
        
        AI-generated content creates a new challenge since it does not
        directly copy existing content but generates new
        text. Traditional methods that identify similarities between a 
        new document and a database of existing documents may fall short of distinguishing AI-generated content from new human work. Large language models aim to  
        create natural, human-like text, making it increasingly
        hard to differentiate generated from human-created text.
        
        Many tools to detect AI-generated text are now offered.
        Some popular tools geared toward educators are 
        Copyleaks AI Content Detector,
        Crossplag, 
        GPTZero,
        Hugging Face OpenAI Detector,
        Originality.ai,
        Turnitin AI Detection and
        ZeroGPT.
        The list of detectors and their capability 
        is constantly changing following the fast-paced changes
        seen in the development of large language models.

        Most tools are based on detecting artifacts 
        of the text generation process, including word choice, 
        writing style, sentence length, and many more. 
        A report by Open AI \cite{solaiman2019release}
        lays out three AI content detection strategies, including a simple 
        classifier learned from scratch, a classifier resulting from fine-tuning an existing language model, or using the probabilities assigned by the model to strings.
        Many existing tools follow the first two approaches. 
        For example, the Hugging Face Open AI detector is a 
        transformer-based classifier that is fine-tuned to detect 
        GPT-2 text. Self-detection introduced in this paper is most closely related 
        to the third approach. However, it does not require access to the model parameters
        to assess probabilities. It relies on the model itself to perform the 
        detection.
        
        
\subsection{Generative AI and Academic Integrity}
       
    Many studies have addressed the ethical implications of AI-generated content in academic contexts in recent years. Notable results can be can be found in \cite{cotton2023chatting}, \cite{liu2023check}, \cite{mitrovic2023chatgpt} and, \cite{anderson2023ai}.

    Gua et al~\cite{guo2023close} introduce the Human ChatGPT Comparison Corpus used to compare ChatGPT compared to human-generated content. They found that part-of-speech (POS) and dependency analysis demonstrate that ChatGPT uses more determination, conjunction, and auxiliary relations, producing longer dependency distances for certain relations. 
    On the other hand, Busch Hausvik~\cite{busch2023too} 
    found that ChatGPT can generate exam answers indistinguishable from human-written text. This raises concerns about academic misconduct. 
    Khalil and Er~\cite{khalil2023will} indicate students could potentially use ChatGPT to bypass plagiarism detection. This indicates that plagiarism detection will need to shift its focus to verifying the origin of the content.

    Yu et al~\cite{yu2023cheat} focus on finding ChatGPT-written content in academic paper abstracts by developing the CHatGPT-writtEn AbsTract (CHEAT) dataset to support the development of detection algorithms.
    Weber-Wulff et al~\cite{weberwulff2023testing}
    compare multiple tools for testing AI-generated text detection, such as Check For AI, Turnitin, ZeroGPT, PlagiarismCheck.
    The results indicate significant limitations in detecting AI-generated content with many false positives and negatives. Detection tools often misclassify AI-generated content as human-written and struggle with obfuscated texts. The conclusion from a study by
    Ventayen~\cite{ventayen2023openai} shows a similar results. These 
    studies show that detecting AI-generated text is a new and
    very difficult problem where new AI models are presented 
    regularly.

\section{AI Self-Detection by Transformer-based Models}

Most detection tools focus on training a classifier that learns to detect artifacts 
introduced by the generative model when generating text. While some types of artifacts may
result from the used base technology, the transformer, many
more will be due to model training, including the chosen training data
and the performed fine-tuning. Since every model can be trained 
differently, creating one detector tool to detect the artifacts
created by all possible generative AI tools is hard to achieve. 

Here, we develop a different approach called self-detection, where we
use the generative model itself to detect its own artifacts to distinguish its own generated text from human-written text. This would
have the advantage that we do not need to learn to detect all generative AI models, but we only need access to a generative AI model for detection. This is a big advantage in a world where new models are
continuously developed and trained.
We start
with an argument about why large language models may have the capability
to detect their own artifacts.

Current large language models use 
the decoder of the transformer architecture 
as their basic building block (see 
\cite{radford2018improving,thoppilan2022lamda,anil2023palm}). 
These models are pre-trained
using the unsupervised task of predicting the next word token on a large 
text corpus. The model learns the following function 
$$P(u_{i+1}|u_{i-k}, ..., u_{i}) = f(u_{i-k}, ..., u_{i}, u_{i+1})$$
to predict the 
probability for each possible next token for the next position $i+1$.
$u_i$ is the $i$-th word token in the sequence and $k$ is the context length.
The model will then predict the token with the highest probability or randomly
choose among the most likely tokens.
During this training phase, the model learns the language's grammar
and acquires facts and knowledge vital to performing well on the
next-word prediction task. The popular Chatbot models are then
fine-tuned (typically using reinforcement learning~\cite{ziegler2020finetuning}) to produce suitable responses to user requests.
An example of this approach is ChatGPT, which is a fine-tuned GPT model\cite{radford2018improving}.

Generating text using a trained model consists of the following steps:

\begin{enumerate}
\item Tokenize the input text.
\item Embed the tokens as numeric vectors and add positional information.
\item Apply multiple transformer blocks using self-attention and
   predict the next token using the transformer's output. 
\item Add the new token to the input sequence and go back to step 2 till 
   a special end token is produced.
\item Convert the generated token sequence back to text.
\end{enumerate}

This approach is autoregressive since it adds one token at a time and the next token 
depends on the previously generated tokens.
The most important innovation of transformers is attention\cite{transformer_paper}, where the model learns to modify tokens to attend to other tokens in the sequence. For example, in the sentence "it is not hot," hot can attend to the word not modifying hot to look more similar to the word cold.

The typical use case for a chatbot is that a user provides the prompt consisting of a request, and the model generates the answer. During the text generation process, the model will attend to the tokens in the prompt to be relevant for the prompt and
to the tokens generated so far to create a consistent answer. This means that
if the complete prompt and the generated text are available, the
model can check if the complete sequence is consistent with its learned function. 

In the following, we will perform several experiments to investigate 
the following hypotheses:

\begin{itemize}
\item [H1:] Generative AI models based on transformers can self-detection
their own generated text.

\item [H2:] Generative AI models based on transformers can self-detect text they have paraphrased.

\item [H3:] Generative AI models cannot detect other model's generated text. 
\end{itemize}

\section{Experimental Setup}

    For the experiments in this paper, we use three models: Open AI's ChatGPT-3.5, 
    Google's Bard and Anthropic’s Claude (all using their September 2023 version).
    We created a new dataset consisting of texts about 50 different topics. We use each model for each topic to generate essays containing approximately 250 words each. The experimental procedure maintained consistency by providing each AI system with an identical prompt 
    (see Appendix~\ref{appendix:prompts}), which instructed them to write an essay based on the given topic. The uniformity in the prompts focused on ensuring the comparability between AI-generated texts in terms of content and length. This process resulted 
    in 50 AI-generated essays produced with a short prompt. 
    Following the initial generation of essays, each original essay underwent a paraphrasing process by the same AI system. 
    We prompted the AI system with the original essay and the instruction to rewrite it (see Appendix~\ref{appendix:prompts}). 
    This procedure resulted in 50 modified versions of those essays. 
    For comparison, we also collected 50 human-written essays of similar length from \url{bbc.com} by manually searching for recent news on the given topics and extracting text passages of about 250 words. Statistics of
    the generated dataset are in Appendix~\ref{appendix:statistics}.
    
    After the creation of the essay dataset, we used zero-shot prompting
    to ask the AI system to perform self-detection.  
    This is a very convenient approach because it can be quickly 
    done with any model and does not require extra steps like 
    fine-tuning.
    We created a new instance of each AI system initiated and posed with a specific query: "If the following text matches its writing pattern and choice of words." The procedure is repeated for the original, paraphrased,
    and human essays, and the results are recorded. We
    also added the result of the AI detection tool ZeroGPT. We do not
    use this result to compare performance but as a baseline to 
    show how challenging the detection task is. 
    The complete dataset with the results of self-detection 
    is available for research at \url{https://github.com/antoniocaiado1/ai-self-detection-study-dataset/}.

    

\section{Results}

\
\begin{table*}[t]
\centering
\begin{tabular}{rlllrrrr}
  \hline
 & Generator & Paraphrased & Detector & Accuracy & PValue (Accuracy $>$ .5) & Detection Rate & Precision \\ 
  \hline
1 & ChatGPT & No & Self-Detection & 0.83 & 0.00 & 0.90 & 0.79 \\ 
  2 & ChatGPT & No & ZeroGPT & 0.66 & 0.00 & 0.64 & 0.67 \\ 
  3 & Bard & No & Self-Detection & 0.94 & 0.00 & 0.96 & 0.92 \\ 
  4 & Bard & No & ZeroGPT & 0.78 & 0.00 & 0.88 & 0.73 \\ 
  5 & Claude & No & Self-Detection & 0.45 & 0.86 & 0.04 & 0.22 \\ 
  6 & Claude & No & ZeroGPT & 0.40 & 0.98 & 0.12 & 0.27 \\ 
  7 & ChatGPT & Yes & Self-Detection & 0.58 & 0.07 & 0.40 & 0.62 \\ 
  8 & ChatGPT & Yes & ZeroGPT & 0.50 & 0.54 & 0.32 & 0.50 \\ 
  9 & Bard & Yes & Self-Detection & 0.92 & 0.00 & 0.92 & 0.92 \\ 
  10 & Bard & Yes & ZeroGPT & 0.72 & 0.00 & 0.76 & 0.70 \\ 
  11 & Claude & Yes & Self-Detection & 0.83 & 0.00 & 0.80 & 0.85 \\ 
  12 & Claude & Yes & ZeroGPT & 0.38 & 0.99 & 0.08 & 0.20 \\ 
   \hline
\end{tabular}
\caption{Results of AI self-detection.}
\label{tab:self_results}
\end{table*}

    \begin{figure*}[t]
        \centering
        \begin{subfigure}{0.45\textwidth}
            \includegraphics[width=\textwidth]{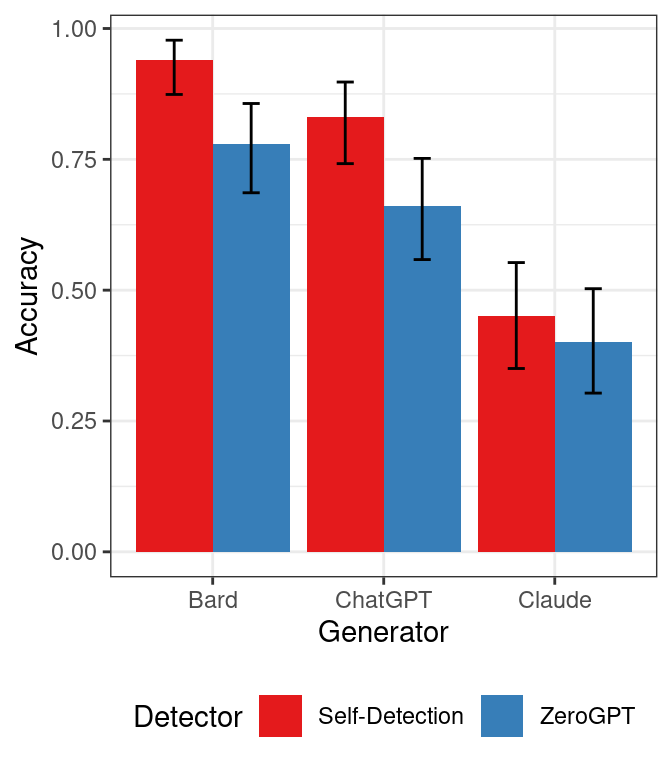}
            \caption{Self-detection accuracy on prompted essays.}
            \label{fig:AI-self_original}
        \end{subfigure}
        \hfill
        \begin{subfigure}{0.45\textwidth}
            \includegraphics[width=\textwidth]{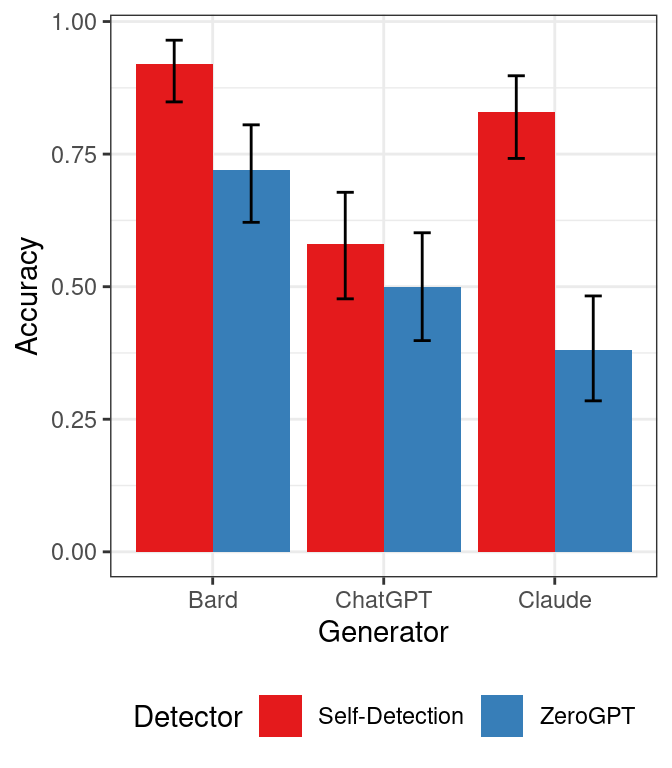}
            \caption{Self-detection accuracy on AI-paraphrased essays.}
            \label{fig:AI-self_paraphrased}
        \end{subfigure}
        \caption{AI self-detection accuracy with 95\% confidence interval.}
        \label{fig:figures}
    \end{figure*}

    For hypotheses H1 and H2, we compare how well
    AI systems can self-detect their own text compared to the 
    human-written texts. Each comparison involves 
    50 AI-generated and 50 human-written texts.
    The results are shown in
    Table~\ref{tab:self_results}. The accuracy results are
    visualized in Figure~\ref{fig:figures}. The charts also include
    error bars indicating the 95\% confidence interval around the
    estimated accuracy. Note that the dataset is always balanced, meaning
    an accuracy of 50\% means random guessing by a model 
    with no detection power. 
    The AI systems show varying abilities to recognize their generated and paraphrased texts.
    
    Hypothesis H1 proposes that generative AI models based on transformers can self-detect their own generated text.
    This can be analyzed using the results in 
    Figure~\ref{fig:AI-self_original}. The chart shows that
    Bard and ChatGPT perform well in 
    distinguishing their generated text from human-written text with
    high accuracy values. The confidence intervals do not span 50\%
    indicating that they can self-detect. Claude, however,
    lacks this ability with a confidence interval spanning an accuracy of
    50\%, so it is not able to self-detect. 
    Note that the ability or inability to self-detect results from two reasons. 
    The ability to self-detect given the transformer approach and
    how well the models mimic human writing. To look into this, we 
    also applied ZeroGPT as a baseline detector. The chosen detector and
    its actual performance are not so important. Still, it is important that ZeroGPT performed much better for text generated
    by Bard and ChatGPT and could not detect Claude's 
    generated text. This may indicate that Claude produced output with 
    harder-to-detect artifacts, which also would make it harder for
    Claude to self-detect.
    
    Hypothesis H2 proposes that generative AI models based on transformers can self-detect text they have paraphrased. The 
    reason for this hypothesis is that the artifacts created by the model should also be present when it rewrites text. However,
    the prompting process differs since it includes the original text, which may lead to different self-detection performances.
    Figure~\ref{fig:AI-self_paraphrased} shows the accuracy of
    paraphrased text versus human-written text.
    The ZeroGPT baseline shows that the performance on the paraphrased
    essays is largely similar to that on the 
    original essays. 
    The results for Bard's self-detection are slightly lower than 
    on the original essays. ChatGPT performs way worse with the 95\% 
    confidence band
    covering 50\%, but Claude seems to be able to self-detect
    its paraphrased content while it was unable to detect its original 
    essay.
    The finding that paraphrasing prevents ChatGPT 
    from self-detecting while increasing Claude's ability to self-detect is very interesting and may be the result of
    the inner workings of these two transformer models. There is
    promise for hypothesis H2. However, the fact that a long 
    prompt and the corresponding attendance values are missing seem
    to make it a far more difficult problem.

\begin{table*}[t]
\centering
\begin{tabular}{rllrrrr}
  \hline
 & Generator & Detector & Accuracy & PValue (Accuracy $>$ .5) & Detection Rate & Precision \\ 
  \hline
1 & ChatGPT & ChatGPT & 0.83 & 0.00 & 0.90 & 0.79 \\ 
  2 & ChatGPT & Bard & 0.48 & 0.69 & 0.04 & 0.33 \\ 
  3 & ChatGPT & Claude & 0.48 & 0.69 & 0.10 & 0.42 \\ 
  4 & Bard & ChatGPT & 0.63 & 0.01 & 0.50 & 0.68 \\ 
  5 & Bard & Bard & 0.94 & 0.00 & 0.96 & 0.92 \\ 
  6 & Bard & Claude & 0.63 & 0.01 & 0.40 & 0.74 \\ 
  7 & Claude & ChatGPT & 0.68 & 0.00 & 0.60 & 0.71 \\ 
  8 & Claude & Bard & 0.46 & 0.82 & 0.00 & 0.00 \\ 
  9 & Claude & Claude & 0.45 & 0.86 & 0.04 & 0.22 \\ 
   \hline
\end{tabular}
\caption{Results of AI self-detection vs. detection by other models.}
\label{tab:cross_results}
\end{table*}
   \begin{figure*}[t]
      \centering
            \includegraphics[width=.45\textwidth]{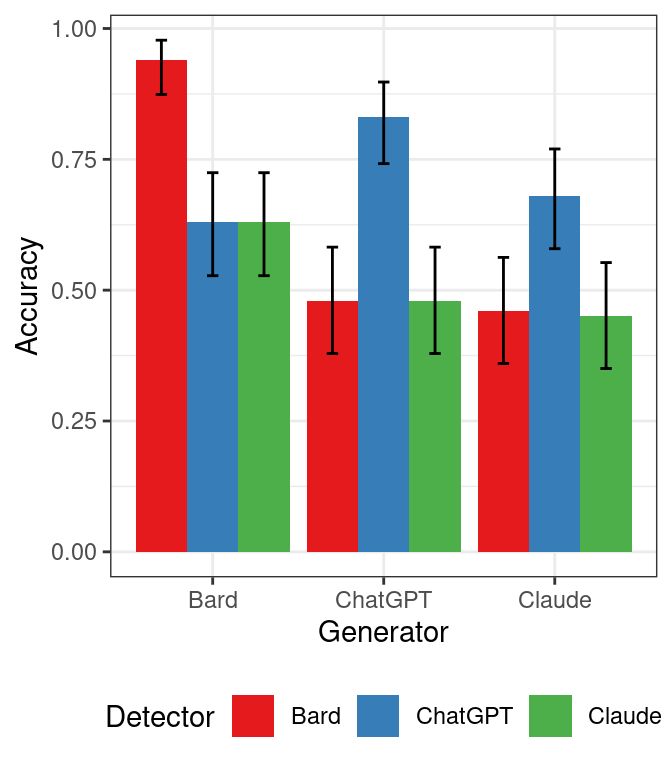}
        \caption{AI self-detection vs. detection by other models accuracy with 95\% confidence interval.}
        \label{fig:AI-cross_detection}
    \end{figure*}
    
     To investigate hypothesis H3, which proposes that AI models cannot detect text 
     generated by other models, we ask each model to determine if the other model's output
     is human-written or AI-generated. The results are shown in 
     Table~\ref{tab:cross_results} and Figure~\ref{fig:AI-cross_detection}.
     We see again that Bard's text is the easiest to detect. Bard's self-detection
     if 94\%. The other model also can detect some of Bard's text, but at 
     a level just above random guessing. ChatGPT can self-detect its  
     generated text, but the other models cannot. For Claude, the situation is very 
     different. Claude cannot self-detect its text, and Bard can also not 
     detect Claude's text, but ChatGPT can detect some of Claude's text.

    If the assumption is correct that self-detection relies on the model's
    knowledge of the model parameters used in the transformer, the H3 should hold. 
    However, the study shows a mixed result for H3. It seems like Bard introduces 
    artifacts that are relatively easy to identify by other models, which also explains
    the good performance of the feature-based AI-content detector ZeroGPT on Bard's 
    output. The other models have no access to Bards's model parameter and, therefore, 
    must also be able to pick up these artifacts. 
    For ChatGPT, H3 seems to apply as expected. Claude's generated text is generally the hardest to detect, which may indicate fewer artifacts. Interestingly, Claude cannot self-detect, but ChatGPT can detect either Claude's artifacts
    or knows Claude's generating model. An explanation could be that either
    ChatGPT or Claude either shared a significant portion of their training sets or  
    could train on each other's generated text. However, this is hard to 
    determine from the outside.
     


    

\section{Discussion}
    Detecting the use of the currently leading AI systems is a difficult task.
    The results in this paper demonstrate varying capabilities of leading AI systems to self-detect their own generated text. Bard performs the best on
    its own work, including originally created essays from a short prompt
    and after paraphrasing a longer given essay.
    ChatGPT performs 
    reasonably well on essays it has created after a short prompt but
    cannot reliably detect essays it has paraphrased. Claude is not 
    able to detect its own created text. 
    This seemingly inconclusive result needs more consideration since it is
    driven by two conflated causes. 

    \begin{enumerate}
        \item The ability of the model to create text with very few 
    detectable artifacts. Since the goal of these systems is to generate 
    human-like text, fewer artifacts that are harder to detect means the model
    gets closer to that goal.
        \item The inherent ability of the model to self-detect can be affected
    by the used architecture, the prompt, and the applied fine-tuning.
    \end{enumerate}

    We use the external AI content detector ZeroGPT to address the first cause.
    ZeroGPT states on its website\footnote{ZeroGPT website: \url{https://zerogpt.cc/}} that it works accurately for
    text created by models including GPT-4, GPT-3, GPT-2, Claude AI, and Google Bard. We use its results as a proxy for how difficult it is
    to detect the text generated by different models. The results 
    in Figure~\ref{fig:AI-self_original}
    show that
    Bard's generated text is the easiest to detect, followed by ChatGPT. Only
    Claude cannot be detected. This indicates that Claude might produce fewer 
    detectable artifacts than the other models. The detection rate of self-detection follows the same trend, indicating that Claude creates 
    text with fewer artifacts, making it harder to distinguish from human writing. Self-detection shows similar detection power compared to 
    ZeroGPT, but note that the goal of this study is not to claim that self-detection is superior to other methods, which would require a large study to compare to many state-of-the-art AI content detection tools. Here, we only investigate the models' basic ability of self-detection.

    In general, the self-detection performance 
    decreases for AI-paraphrased text
    (shown in Figure~\ref{fig:AI-self_paraphrased}). This may be affected by
    the inherent ability of the transformer-based models to self-detect. 
    An important part of why transformer-based large language models 
    process the prompt and generate text so successfully is using the attention mechanism. Attention allows the model to learn how to modify
    tokens based on previously seen tokens to include context information
    before it uses a learned function to predict the next word. 
    Since the transformer has access to its own attention mechanism and the 
    prediction function, we have 
    hypothesized that transformer-based generative AI models can self-detect their own generated text. An important issue is that the prompt text is available during text generation and is included in the attention calculation. The used prompt is typically not available during self-detection. This means that the attention to the tokens in the prompt cannot be calculated, reducing the ability to self-detect. 
    A counter-intuitive finding is that Claude has difficulties in self-detecting its originally generated content but can detect
    content that it has paraphrased with a high degree of accuracy while
    the baseline detector still cannot detect it.

    This initial study has several important limitations.
    \begin{itemize}
        \item This study is limited by the small dataset containing a randomized set of topics and a simplified paraphrasing approach.
        \item This experiment only utilizes three popular AI systems---ChatGPT, BARD, and Claude.
        \item Generative AI systems are constantly evolving, and the 
        systems are changing quickly (e.g., by training on additional data, changes in pre-prompts, and changes in the used architecture). This makes comparisons difficult, and detailed results may quickly become irrelevant.
        \item Only a single conventional plagiarism detection tool, ZeroGPT, has been used as a baseline to reason about the artifacts present in the output of different models. Many other popular AI content detection tools exist (Turnitin, PlagiarismCheck, GPT Zero, etc.).
    \end{itemize} 
    

    While AI content detection tools have the advantage that they
    can be trained to identify the artifacts of multiple generative AI tools,
    they need to be updated to add detection capabilities for a new model or 
    when models change.
    A significant disadvantage is that
    self-detection can only detect its own work by using knowledge of its 
    generation process and the artifacts that it creates. However, in a world 
    where new models are introduced at a break-neck pace, it may be easier and faster to add this new model to the set of models that are asked to self-detect instead of creating a large amount of data with the models and retraining a standard AI content detector.
\section{Conclusion}
    Detecting AI-generated content, which includes proper attribution of authorship and addressing questions of remuneration of the creator of the content used to train these models, is becoming increasingly important for many applications. Especially in academia, generative AI has many uses that can improve
    learning by generating explanations for students, but it can also 
    detract from
    learning by enabling students to let AI solve their exercises.  

    
    This study's unique contribution lies in introducing self-detection, a step forward in addressing the challenges posed by AI systems. We describe why
    transformer-based systems should have the capability to self-detect and 
    demonstrate this capability in a first small study. We identify the main
    limitation of self-detection as the unavailability of the original prompt.

    %

    The presented first study is very limited. Here are some topics
    to explore in future studies.
    \begin{itemize}
        \item Use a larger dataset with more diverse generated text.
        \item Explore more different generative AI models.
        \item Compare the performance of self-detection with the currently
        best detectors.
        \item Explore how prompt engineering affects self-detection. For example,
        use few-shot prompting for self-detection.
    \end{itemize}

\bibliographystyle{IEEEtran}
\bibliography{Sections/references}
\balance

\clearpage
\onecolumn
\appendices



\section{Prompts used in this Study}
    \label{appendix:prompts}

    \begin{table}[!h]
        \centering
        \begin{tabular}{p{3.5 cm}p{12 cm}}
            \hline 
             Task  &  Prompt  \\
            \hline
          Essay Generation & Write an essay within 250 words regarding (topic name) in one paragraph \\[2mm]
          
          Paraphrased Essay Generation  & Paraphrase the following paragraph: (previously generated paragraph)\\[2mm]
          
           AI Self-Detection  & Check if the following paragraph matches your text patterns and choice of words for generating the response. If it matches, respond \textbf{TRUE}; otherwise, \textbf{FALSE}.\\[2mm]
           
             AI Detecting other AI Content  & Check if the given paragraph matches or contains AI jargon or Human written patterns. Give your answer as either \textbf{Human} or \textbf{AI}.\\
            \hline  
        \end{tabular}
        \caption{Prompts used in this study}
        \label{tab:prompt}
    \end{table}

\section{Statistics of the Data Set}
    \label{appendix:statistics}

\begin{table}[h]
\centering
\begin{tabular}{rlllrrrrr}
  \hline
 & Source & Model & Paraphrased & Number of Essays & Word Count & Sentence Count & Words/Sentence & Newline Count \\ 
  \hline
1 & AI & ChatGPT & No &  50 & 193.90 & 7.98 & 24.30 & 0.00 \\ 
  2 & AI & ChatGPT & Yes &  50 & 194.36 & 8.10 & 24.00 & 0.00 \\ 
  3 & AI & Bard & No &  50 & 292.30 & 6.06 & 48.23 & 10.86 \\ 
  4 & AI & Bard & Yes &  50 & 251.48 & 5.24 & 47.99 & 9.32 \\ 
  5 & AI & Claude & No &  50 & 206.38 & 11.76 & 17.55 & 0.00 \\ 
  6 & AI & Claude & Yes &  50 & 187.40 & 11.06 & 16.94 & 0.00 \\ 
  7 & Human & None &  &  50 & 348.26 & 11.04 & 31.55 & 6.24 \\ 
   \hline
\end{tabular}
\caption{Average statistics for different groups in the dataset}
\label{tab:stats}
\end{table}

\clearpage

\section{Example Data}
The complete dataset can be found at \url{https://github.com/antoniocaiado1/ai-self-detection-study-dataset/}.

\subsection{ChatGPT's Essay Identification Data}
    \begin{table}[!h]
        \centering
        \begin{tabular}{p{2cm}p{6.5cm}p{6.5cm}}
        \hline
             Topic  &  Generated Answer  &  Paraphrased Answer   \\
            \hline
             James Webb Telescope  & The James Webb Space Telescope (JWST) stands as a monumental advancement in humanity's quest to unravel the mysteries of the universe. Scheduled for launch, this successor to the Hubble Space Telescope promises to revolutionize our understanding of celestial phenomena and deepen our insights into the cosmos. Designed collaboratively by NASA, the European Space Agency (ESA), and the Canadian Space Agency (CSA), the JWST boasts an array of cutting-edge technologies that enable it to observe the universe in ways previously unattainable. Operating primarily in the infrared spectrum, the telescope's intricate instruments will allow it to peer through cosmic dust clouds, capture the glow of distant galaxies, and study the atmospheres of exoplanets. Moreover, the JWST's colossal sunshield will protect its sensitive instruments from the Sun's heat and light, ensuring precise and clear observations. By probing the universe's earliest moments, studying the formation of stars and galaxies, the JWST is poised to provide unprecedented insights into the fundamental questions surrounding our existence and the cosmos. & The James Webb Space Telescope (JWST) represents a significant leap in humanity's quest to comprehend the universe's enigmas. With a launch date on the horizon, this successor to the Hubble Space Telescope pledges to transform our grasp of celestial occurrences and deepen our cosmic insights. Collaboratively crafted by NASA, the European Space Agency (ESA), and the Canadian Space Agency (CSA), the JWST showcases an array of state-of-the-art technologies, enabling it to observe the cosmos in unprecedented ways. Primarily functioning in the infrared range, the telescope's intricate tools will facilitate peering through interstellar dust, capturing remote galaxy glows, and scrutinizing exoplanet atmospheres. Additionally, the JWST's expansive sunshield will safeguard its delicate equipment from solar heat and brilliance, ensuring meticulous and lucid observations. By investigating the universe's infancy, examining star and galaxy formation, the JWST is poised to deliver unparalleled revelations into the fundamental queries surrounding human existence and the wider cosmos.\\[5mm]

             Global Warming   & Global warming, an alarming consequence of human activities, has emerged as a pressing threat to our planet. The unrestrained burning of fossil fuels, deforestation, and industrial processes have unleashed an influx of greenhouse gases into the atmosphere, intensifying the natural greenhouse effect and causing Earth's temperatures to rise. Consequently, ice sheets and glaciers are melting at an accelerated pace, fueling sea level rise and endangering coastal regions. The frequency and intensity of extreme weather events have surged, resulting in devastating hurricanes, prolonged droughts, and erratic rainfall patterns. These changes disrupt ecosystems, endangering various species and diminishing biodiversity. Urgent global action is imperative to curb this crisis. Governments, industries, and individuals must transition to sustainable energy sources, enhance forest conservation efforts, and adopt eco-friendly practices. Furthermore, raising awareness about the impacts of global warming can drive collective efforts to reduce carbon emissions and promote environmental stewardship. By recognizing the gravity of the situation and taking comprehensive measures, we can mitigate global warming's adverse effects and secure a sustainable future for the generations to come. & Global warming, a concerning outcome of human actions, has arisen as an imminent menace to our planet. Uncontrolled combustion of fossil fuels, deforestation, and industrial operations have released a surge of greenhouse gases into the atmosphere, intensifying the innate greenhouse effect and leading to a rise in Earth's temperatures. This has led to the accelerated melting of ice sheets and glaciers, contributing to the elevation of sea levels and jeopardizing coastal areas. The frequency and severity of extreme weather occurrences have risen substantially, causing catastrophic hurricanes, prolonged periods of drought, and irregular patterns of precipitation. These alterations disrupt ecosystems, imperil diverse species, and lessen biodiversity. Swift worldwide measures are vital to alleviate this predicament. Governments, industries, and individuals should transition toward sustainable energy sources, amplify endeavors for forest preservation, and embrace environmentally conscious practices. Moreover, heightening awareness about the repercussions of global warming can stimulate collective endeavors to decrease carbon emissions and foster ecological guardianship. By acknowledging the seriousness of the situation and implementing comprehensive steps, we can mitigate the detrimental repercussions of global warming and ensure a sustainable future for future generations.   \\[2mm]
       
            \hline
        \end{tabular}
        \caption{ChatGPT's Essay Identification Data}
        \label{tab:ChatGPT}
    \end{table}

\clearpage

\subsection{BARD's Essay Identification Data}
    \begin{table}[!h]
        \centering
        \begin{tabular}{p{2cm}p{6.5cm}p{6.5cm}}
        \hline
             Topic  &  Generated Answer  &  Paraphrased Answer    \\
            \hline
             James Webb Telescope  & The James Webb Space Telescope (JWST) is the most powerful telescope ever built. It was launched on December 25, 2021, and is currently orbiting the Sun about a million miles away from Earth. JWST is designed to study the infrared universe, which allows it to see objects that are too old, distant, or faint for other telescopes to see. JWST's primary mirror is 6.5 meters in diameter, which is about six times larger than the Hubble Space Telescope's mirror. This larger mirror allows JWST to collect more light and see fainter objects. JWST also has four instruments that are specifically designed to study the infrared universe. JWST is also expected to provide new insights into the formation of stars and galaxies, and the evolution of the universe. The James Webb Space Telescope is a truly remarkable telescope that has the potential to revolutionize our understanding of the universe. It is a testament to the ingenuity and perseverance of the scientists and engineers who built it.
            & The James Webb Space Telescope is a powerful telescope that can see objects that are too old, distant, or faint for other telescopes to see. It was launched in 2021 and is currently orbiting the Sun about a million miles away from Earth. JWST is expected to make groundbreaking discoveries about the universe, including seeing the first stars and galaxies that formed after the Big Bang. JWST's primary mirror is 6.5 meters in diameter, which is about six times larger than the Hubble Space Telescope's mirror. This larger mirror allows JWST to collect more light and see fainter objects. JWST also has four instruments that are specifically designed to study the infrared universe. The James Webb Space Telescope is a truly remarkable telescope that has the potential to revolutionize our understanding of the universe. It is a testament to the ingenuity and perseverance of the scientists and engineers who built it.   \\[5mm]
          
            Global Warming & Global warming is the long-term heating of Earth's climate system observed since the pre-industrial period (between 1850 and 1900) due to human activities, primarily fossil fuel burning, which increases heat-trapping greenhouse gas levels in Earth's atmosphere. The term is frequently used interchangeably with the term climate change, though the latter refers to both human- and naturally produced warming and the effects it has on our planet. The main cause of global warming is the release of greenhouse gases into the atmosphere. Greenhouse gases trap heat from the sun, causing the planet to warm. The primary greenhouse gases are carbon dioxide, methane, nitrous oxide, and ozone.The effects of global warming are already being felt around the world. These effects include rising sea levels, melting glaciers, more extreme weather events, and changes in plant and animal life. Global warming is a serious threat to our planet and its inhabitants. We need to take urgent action to reduce greenhouse gas emissions and mitigate the effects of climate change. & Global warming is the long-term heating of Earth's climate system. It is caused by human activities, such as burning fossil fuels, which release greenhouse gases into the atmosphere. These gases trap heat, causing the planet to warm.The effects of global warming are already being felt around the world. Sea levels are rising, glaciers are melting, and extreme weather events are becoming more common. Global warming is a serious threat to our planet and its inhabitants.We need to take action to reduce greenhouse gas emissions and mitigate the effects of climate change. This includes switching to renewable energy sources, improving energy efficiency, and reducing deforestation. Individuals can also make a difference by making changes in their own lives, such as driving less, using less energy, and eating less meat.By taking action now, we can help to protect our planet for future generations. \\
            \hline

        \end{tabular}
        \caption{BARD's Essay Identification Data}
        \label{tab:BARD}
    \end{table}

\clearpage

\subsection{Claude's Essay Identification Data}
    \begin{table}[!h]
        \centering
        \begin{tabular}{p{2cm}p{6.6cm}p{6.5cm}}
        \hline
             Topic  &  Generated Answer  &  Paraphrased Answer   \\
            \hline
             James Webb Telescope  & The James Webb Space Telescope (JWST) is a large, infrared-optimized space telescope launched by NASA, ESA, and the Canadian Space Agency in 2021. JWST is the most powerful space telescope ever built, acting as the successor to the Hubble Space Telescope. Stationed nearly 1 million miles from Earth at the second Lagrange point, JWST utilizes a 6.5-meter primary mirror and four scientific instruments to observe extremely distant celestial objects. Its specialized infrared technology allows JWST to penetrate cosmic dust clouds and detect the faint light from the earliest stars and galaxies in the universe over 13.5 billion years ago. One of JWST’s primary scientific goals is observing these primordial galaxies during the reionization era to understand the origins of the universe. JWST will also analyze the atmospheres of exoplanets to search for the potential chemical indications of life outside our solar system. The precise launch and orbital insertion of JWST represent major achievements in space engineering and technology. With a minimum lifetime of 5-10 years, JWST is expected to revolutionize many areas of astronomy by revealing unprecedented views of the infrared universe. 
            & The James Webb Space Telescope, launched in 2021, is the most powerful space telescope ever built. Stationed far from Earth, it uses specialized infrared technology to observe extremely distant cosmic objects, allowing it to see farther back in time than any previous telescope. As Hubble's successor, one of Webb's main goals is studying the earliest galaxies that emerged over 13 billion years ago right after the Big Bang, to understand the origins of the universe. It will also analyze atmospheres of exoplanets orbiting other stars to potentially find signs of life outside our solar system. The precise engineering and technology behind Webb's successful launch and deployment mark major space achievements. With at least a 5-10 year operational lifetime, Webb is expected to revolutionize astronomy by unveiling unprecedented views of the early infrared universe. \\[5mm]
           
            Global Warming & Global warming is the steady rise in Earth's average temperature over the past century primarily due to the burning of fossil fuels like coal, oil and gas. This human-caused climate change is leading to alterations in global weather patterns, melting glaciers and polar ice caps, and rising sea levels. Unless drastic action is taken, these trends will continue and intensify in the coming decades. The consequences of unchecked global warming could be catastrophic, including more extreme weather like droughts, floods and storms, the spread of diseases, reduced agricultural yields leading to food shortages, and the destruction of ecosystems unable to adapt quickly enough. To prevent the most severe impacts, the world must transition from fossil fuels to renewable energy sources like solar and wind power while also practicing conservation and reforestation. Individuals can help by making lifestyle changes to reduce their carbon footprint, but systemic change by governments and corporations will be necessary for major progress. Urgent action is required to curb greenhouse gas emissions and limit further warming to 1.5 degrees Celsius above pre-industrial levels, the target set in the Paris Climate Agreement to avoid environmental tipping points leading to runaway climate change. With global cooperation, it may still be possible to create a sustainable future, but the window for effective action is closing rapidly. & The planet is heating up at an alarming rate largely due to human activities like burning fossil fuels. This is causing major shifts in weather patterns, melting of ice sheets, and rising oceans. If we stay on the current path, the future impacts could be devastating - more extreme weather events, spreading of diseases, food shortages, ecosystem destruction, and so on. To avoid catastrophic climate change, we need to quickly transition our energy systems to renewables like solar and wind while also conserving resources and replanting forests. Individuals can help by reducing their carbon footprint, but large-scale change by governments and companies will be essential for real progress. There is an urgent need to cut emissions and limit additional warming to 1.5°C over pre-industrial levels, the target of the Paris Agreement, to avoid triggering feedback loops that could spiral out of control. With global teamwork, it may still be possible to create a sustainable world, but time is running out fast. Decisive action now is critical. \\
            \hline

        \end{tabular}
        \caption{Claude's Essay Identification Data}
        \label{tab:Claude}
    \end{table}

\clearpage

\section{AI Systems' Human Written Text Identification Data}
In this particular task, the AI systems were asked to write an essay about a particular topic and paraphrase it. Later, it was checked whether the system could detect its own generation. For this purpose, the same prompts, shown in \ref{tab:prompt}, were given in every iteration. 
    \begin{table}[!h]
        \centering
        \begin{tabular}{p{3cm}p{12cm}}
        \hline
             Topic  &  Human Text \\
            \hline
             Global Warming   & Climate change is already shaping what the future will look like and plunging the world into crisis. Cities are adapting to more frequent and intense extreme weather events, like superstorms and heatwaves. People are already battling more destructive wildfires, salvaging flooded homes, or migrating to escape sea level rise. Policies and economies are also changing as world leaders and businesses try to cut down global greenhouse gas emissions. How energy is produced is shifting, too — from fossil fuels to carbon-free renewable alternatives like solar and wind power. New technologies, from next-generation nuclear energy to devices that capture carbon from the atmosphere, are in development as potential solutions. The Verge is following it all as the world reckons with the climate crisis.  \\[5mm]
            
             James Webb Telescope   & NASA’s James Webb Space Telescope’s high resolution, near-infrared look at Herbig-Haro 211 reveals exquisite detail of the outflow of a young star, an infantile analogue of our Sun. Herbig-Haro objects are formed when stellar winds or jets of gas spewing from newborn stars form shock waves colliding with nearby gas and dust at high speeds. The image showcases a series of bow shocks to the southeast (lower-left) and northwest (upper-right) as well as the narrow bipolar jet that powers them in unprecedented detail. Molecules excited by the turbulent conditions, including molecular hydrogen, carbon monoxide and silicon monoxide, emit infrared light, collected by Webb, that map out the structure of the outflows. \\[5mm]
            
             Genetic Engineering   & Religion provides the strongest grounds for protesting genetic engineering. So it is not surprising that most of the resistance to all new reproductive technologies comes from people with religious beliefs. This resistance is deeply rooted in fundamental religious norms. According to the Judeo-Christian tradition, humans were created in the “image” and “likeness” of God (Genesis 1:26-27), which, according to some interpreters, means both the given nature of man and their perfection, the goal towards which they must strive; and from the point of view of others, “image” and “likeness” are synonymous. Humans are likened to God, first of all, in that they were given power over nature (Ps. 8), and also in that they received from the Creator the “breath of life.” Thanks to this, a person becomes a “living soul.” This concept means a living personality, the unity of vital forces, the “I” of a person. Soul and flesh are characterized by organic unity (in contrast to the Greek philosophical dualism, which contrasted spirit and flesh). Some people believe genetic engineering is morally wrong because it interferes with God’s plan for humanity. They believe that we are playing with fire by altering the genes of living organisms and that this could have catastrophic consequences for both humans and the environment. \\
            \hline
        \end{tabular}
        \caption{Human Written Essay Identification Data}
        \label{tab:Claude}
\end{table}



\end{document}